\documentclass[times,twocolumn,final,authoryear]{elsarticle}

\usepackage{ycviu}
\usepackage{times}
\usepackage{epsfig}
\usepackage{graphicx}
\usepackage{amsmath}
\usepackage{amssymb}
\usepackage{epsfig}
\usepackage{amsmath}
\usepackage{amssymb}
\usepackage{algorithm}
\usepackage{algorithm,algorithmicx,algpseudocode}
\usepackage{graphicx}
\usepackage{subcaption}
\usepackage{multirow}
\usepackage{verbatim}
\usepackage{framed,multirow}
\usepackage{comment}
 \usepackage{multirow}
\usepackage[normalem]{ulem}
\useunder{\uline}{\ul}{}
\usepackage{multirow}
\usepackage[normalem]{ulem}
\useunder{\uline}{\ul}{}
\usepackage{multirow}

\usepackage{amssymb}
\usepackage{latexsym}

\usepackage{url}
\usepackage{xcolor}
\definecolor{newcolor}{rgb}{.8,.349,.1}

\journal{Computer Vision and Image Understanding}

\begin{document}

\begin{frontmatter}

%\corref{cor2}
%\cortext[cor2]{Corresponding author: 
%  Tel.: +0-000-000-0000;  
%  fax: +0-000-000-0000;}
%\ead{author@author.com}

\title{MTCD: Cataract Detection via Near Infrared Eye Images}

\author[1]{Pavani \snm{Tripathi}} 
\author[2]{Yasmeena \snm{Akhter}}
\author[2]{Mahapara \snm{Khurshid}}
\author[1]{Aditya \snm{Lakra}}
\author[1]{Rohit \snm{Keshari}}
\author[2]{Mayank \snm{Vatsa}}
\author[2]{Richa \snm{Singh}}

\address[1]{IIIT-Delhi, New Delhi, India}
\address[2]{IIT Jodhpur, Rajasthan, India}

\begin{abstract}
Globally, cataract is a common eye disease and one of the leading causes of blindness and vision impairment. The traditional process of detecting cataracts involves eye examination using a slit-lamp microscope or ophthalmoscope by an ophthalmologist, who checks for clouding of the normally clear lens of the eye. The lack of resources and unavailability of a sufficient number of experts pose a burden to the healthcare system throughout the world, and researchers are exploring the use of AI solutions for assisting the experts. Inspired by the progress in iris recognition, in this research, we present a novel algorithm for cataract detection using near-infrared eye images. The NIR cameras, which are popularly used in iris recognition, are of relatively low cost and easy to operate compared to ophthalmoscope setup for data capture. However, such NIR images have not been explored for cataract detection.  We present deep learning-based eye segmentation and multitask network classification networks for cataract detection using NIR images as input.  The proposed segmentation algorithm efficiently and effectively detects non-ideal eye boundaries and is cost-effective, and the classification network yields very high classification performance on the cataract dataset. 
\end{abstract}


\end{frontmatter}

\section{ Introduction}

\begin{comment}
\begin{figure}[!h]
\centering
\includegraphics[width=3.2in]{fig/ero_dilation.png}
\caption{Showcasing the challenges faced during segmentation}
\label{fig:cataract_samples}
\end{figure}

\end{comment}

Cataract is an age-related ocular disorder in which the eye lens develops a cloudy layer due to the breaking down of proteins in the eye, which makes it opaque, leading to blurry vision. Both eyes of a person can have a different level of cataract and can develop at the different or same time. It is one of the most common eye diseases and is one of the primary causes of blindness \citep{pascolini2012global}. According to the \textit{National Blindness and Visual Impairment Survey of India 2015-19}, people above the age of 50 years may develop blindness due to cataract. The condition contributes to the $66.2\%$ blindness cases, $80.7\%$ of severe visual impairment cases, and $70.2\%$ moderate visual impairment cases in this age group. According to \cite{india}, in India, 50-80\% of bilateral blindness cases can be attributed to cataract. These numbers demonstrate the need of detecting and correcting cataract in time. 
\begin{figure}[!t]
\centering
\includegraphics[width=2.45in]{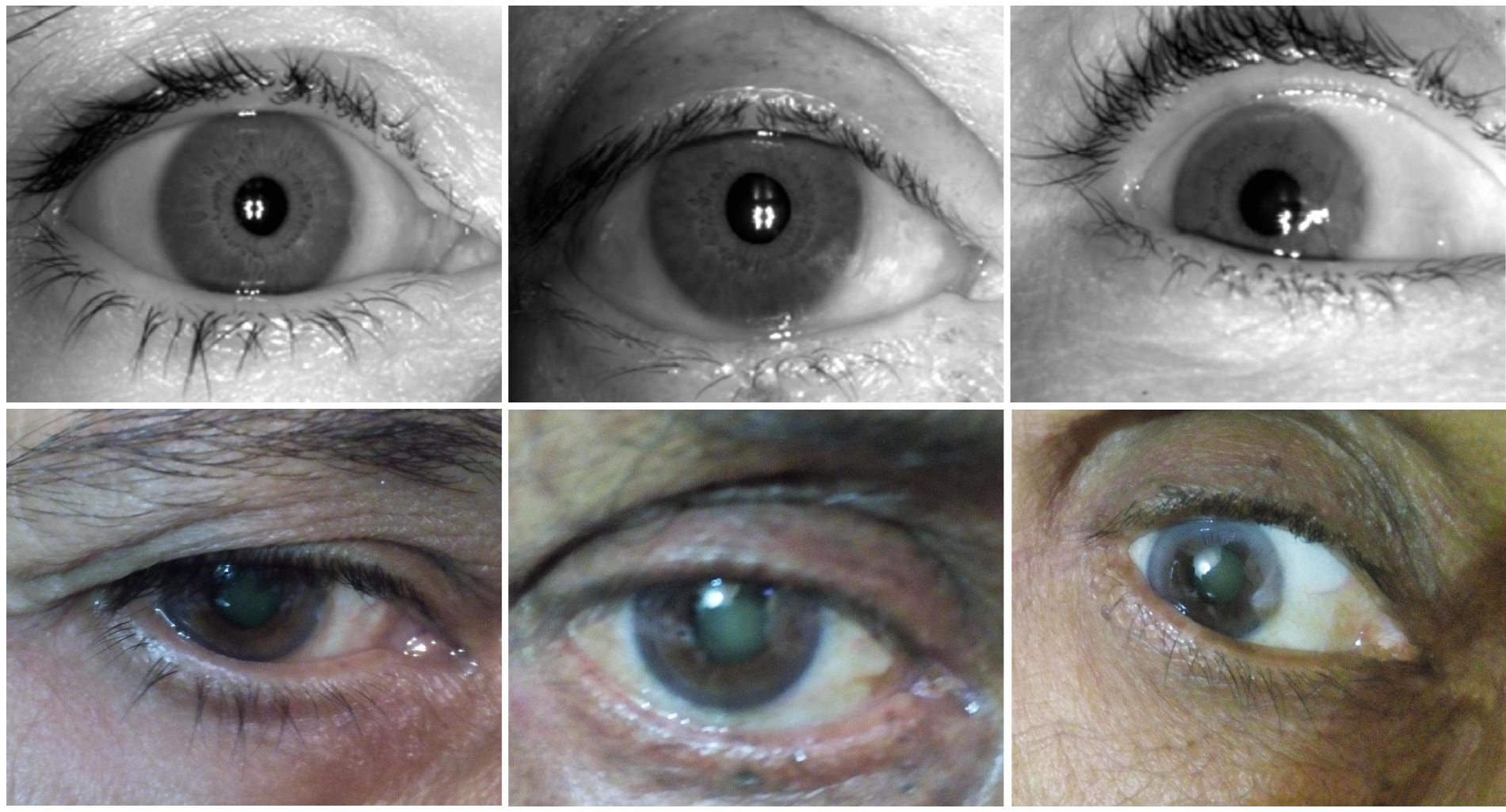}
\caption{Showcasing the affected samples of pre and post cataract from NIR spectrum(top row) and visible spectrum (bottom row)}
\label{fig:catract_samples}
\end{figure}

%as shown in Fig. \ref{fig:tradvsprop}(a) \footnote{https://www.aao.org/eye-health/diseases/what-are-cataracts}
The current process for cataract detection involves using a slit-lamp or an ophthalmoscope for capturing the eye images, and an ophthalmologist examines and tests the eyes of the patient to diagnose the presence of a cataract. While this is the \textit{gold standard}, the rate of blindness, particularly in remote rural areas, is more than the trained ophthalmologists and resources~\citep{murthy2008current}. On the other hand, for biometrics authentication, the low cost near infra-red (NIR) cameras are used in iris recognition. These cameras are available in different form factors, i.e. single eye and dual eye, and they are easy to use. We postulate that eye images obtained from these cameras can help design low cost, accessible, and easy-to-use solutions for cataract detection. 

\begin{figure}
\centering
\includegraphics[width=3.1in]{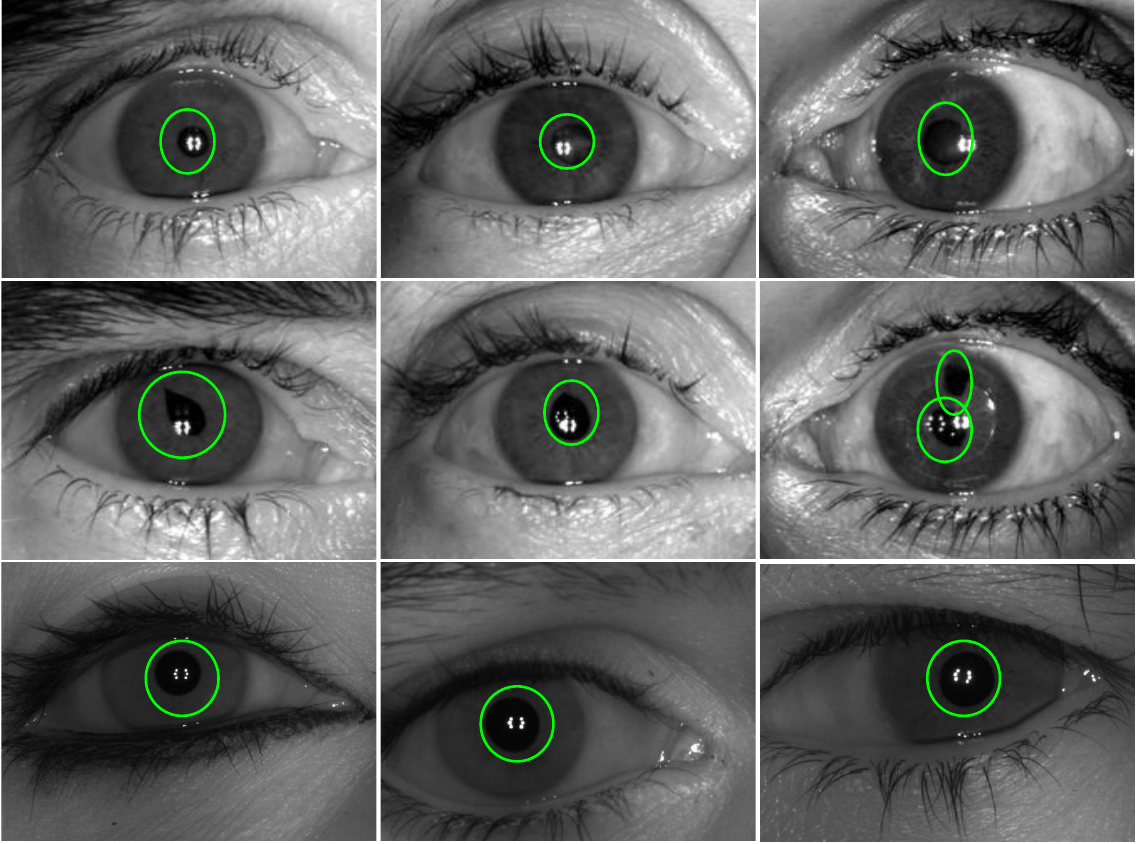}
\caption{Showcasing the visual differences in pupil and iris in the pre and post cataract samples. Top row shows the images with cataract and before surgery. Bottom row shows images with cataract removed after surgery.}
\label{fig:Cataract_pre_post}
\end{figure}

 As shown in Fig. \ref{fig:catract_samples}, NIR eye images provide iris and pupil region which can be utilized to explore  whether these images are useful for cataract detection. However, these samples also highlight the challenges involved in designing an automated algorithm. textcolor{blue}{As shown in  Fig. \ref{fig:Cataract_pre_post}}, the captured images may not be ideal because of: (i) drooping eyelids due to old age, (ii) inadequate camera-to-eye distance and angle, and (iii) excessive contraction or dilation of pupil due to other medical conditions (or ongoing medications such as blood thinner). Therefore, as the first step, designing an efficient segmentation algorithm that segments the iris and pupil regions from the acquired non-ideal images is important. Once the iris and pupil are segmented, the proposed approach involves designing the feature extraction and classification algorithm to differentiate between healthy eye images and images with cataract. In the feature extraction and classification stage as well, the primary challenges are irregular shape and size of iris and pupil. Depending on the kind of occlusion present in the eye image, the angle of capture, and the shape of iris/pupil, the segmented iris and pupil regions can be of different shapes. The classification algorithm should account for these variabilities and perform accurate classification.  
 %%%%%%%%%%%%%%%%

\begin{figure}
\centering
\includegraphics[width=3.375in]{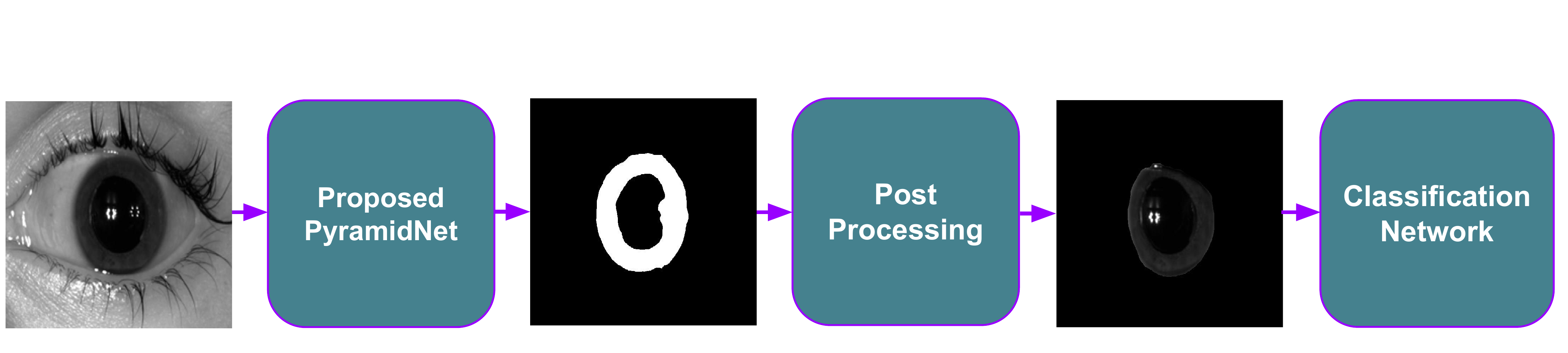}
\caption{Proposed Pipeline of the MTCD Approach (Best viewed in colour). Architecture of Segmentation Network and Classification Network are shown in Fig. \ref{fig:fig_arch} and Fig. \ref{fig:Class}, respectively.}
\label{fig:pipeline}
\end{figure}

To address the above-mentioned challenges, this research presents an automated algorithm, termed as MTCD, for cataract detection from NIR images. As shown in Fig.\ref{fig:pipeline}, the input NIR eye image is first processed using the proposed hierarchical pyramid network termed as $PyramidNet$ to segment iris and pupil patterns from images of eyes acquired in unconstrained environments, in the presence of five different covariates, viz. at-a-distance, clouding for pupil due to cataract, punctured iris due to cataract surgery, and excessive contraction or dilation. After post-processing, the segmented eye image (with iris and pupil boundaries) are then used by a multitask deep learning approach that performs two tasks: the first task classifies the image as \textit{healthy} or \textit{unhealthy}, and the second task classifies the images to one of three classes: \textit{pre-cataract}, \textit{post-cataract}, and \textit{others}. The class 'others' consists of samples that are neither suffering from cataract nor have undergone surgery. The results of the proposed cataract detection algorithm are demonstrated on the publicly available IIITD Cataract Surgery dataset \citep{nigamphacoemulsification}. We further evaluate the performance of the proposed $PyramidNet$ on four challenging eye (iris) datasets that comprise the covariates mentioned above. Since cataract is assessed in the presence of eye drops used for dilating the pupil, we have also prepared a Pupil Dilation dataset comprising images before and after the use of eye drops\footnote{To the best of our knowledge, this is the only dataset in the research community and it will be released to the research community.}. The results on different datasets show that the proposed algorithm yields the best performance in terms of both computation efficiency and memory requirements.

\section{Related Work}

Literature review is divided into two subsections: (i) cataract related and (ii) iris and pupil segmentation related.
\subsection{Cataract Related}
There are limited efforts in automating cataract detection. ~\cite{srivastava2014automatic} have proposed a method to grade the nuclear cataract slit-lamp images using gray level image gradients. \cite{YANG201645} used an ensemble approach on models by exploiting three independent feature sets; wavelet, sketch, and texture-based features for grading the cataract fundus images. \cite{ran2018cataract} have extracted features from fundus images using a three-layer deep convolution neural network (CNN) and random forests (RF) to grade the cataract. They have demonstrated that RF improves grading accuracy. \cite{pratap2019computer}  have used pre-trained AlexNet for feature extraction for fundus image and support vector machines (SVM) for classifying images in different categories of cataract.

\cite{ZHANG2019104978} have implemented a framework to grade the cataract into six levels using feature fusion approach obtained via ResNet18 model and handcrafted GLCM features. \cite{xu2019fully} aimed at grading cataracts from slit-lamp photos using Faster-RCNN to locate the nuclear region and finally used ResNet101 to grade the samples.
%~\cite{pratap2019computer} have proposed a transfer learning-based approach for cataract detection in fundus images into different grades such as normal, mild, moderate, and severe. Pretrained AlexNet model is used to extract features, and finally, SVM is used to perform classification.
\cite{xu2019hybrid} have used the deep model to learn useful features directly from input fundus images for grading the cataract and employed the deconvolution network method to investigate how CNN characterizes cataract layer-by-layer. %Authors found including the local vascular information with global features extracted from fundus image plays a crucial role for this desired problem. 
\cite{grammatikopoulou2019cadis} have proposed an approach for semantic segmentation in cataract surgery videos.
\cite{9283218} proposed GraNet, a CNN-based model, by introducing a point-wise convolution method to learn high-level features for the classification of nuclear cataract from anterior segment optical coherence tomography (AS-OCT) images.
To the best of our knowledge, no work has been reported which utilizes images acquired in the NIR spectrum. The proposed work aims to use NIR eye images as the input to cataract classification. 

\subsection{Iris and Pupil Segmentation Related}

The literature on iris and pupil segmentation in iris biometrics is very rich. Starting with pre-deep learning approaches such as \cite{daugman1993high}, \cite{vatsa2008improving,zhang2010texture} to learning-based approaches \citep{ICCV_2015, radman2017automated}, most of the algorithms focus on near-ideal eye imaging. In the recent literature, Convolutional Neural Networks (CNNs) based approaches are more prevalent. These approaches provide an end-to-end mechanism to search for optimal iris and pupil boundaries. Since segmentation requires the model to correctly segment very fine regions such as iris pixels occluded by eyelashes and specular reflections present in the pupil or iris, the targeted models are designed for segmentation in non-cooperative scenarios \citep{liu_MFCN,arsalan2017deep,lakra2018segdensenet,hofbauer2019exploiting,hu2019icb,wang2020towards}. 
To the best of our knowledge, there is no segmentation algorithm designed for eye images affected due to cataract or post-cataract surgery. Existing algorithms which work well on normal eyes but may not work properly due to the artifacts due to cloudy pupil (or any other medical conditions), small punctures or irregularities in iris that may have resulted due to cataract surgery, or pupil may be medically dilated. In this research, one of the contributions is proposing a segmentation algorithm that helps to segment iris and pupil boundaries in medically affected eyes. 
\begin{figure*}[!t]
\centering
\includegraphics[width=7.2in]{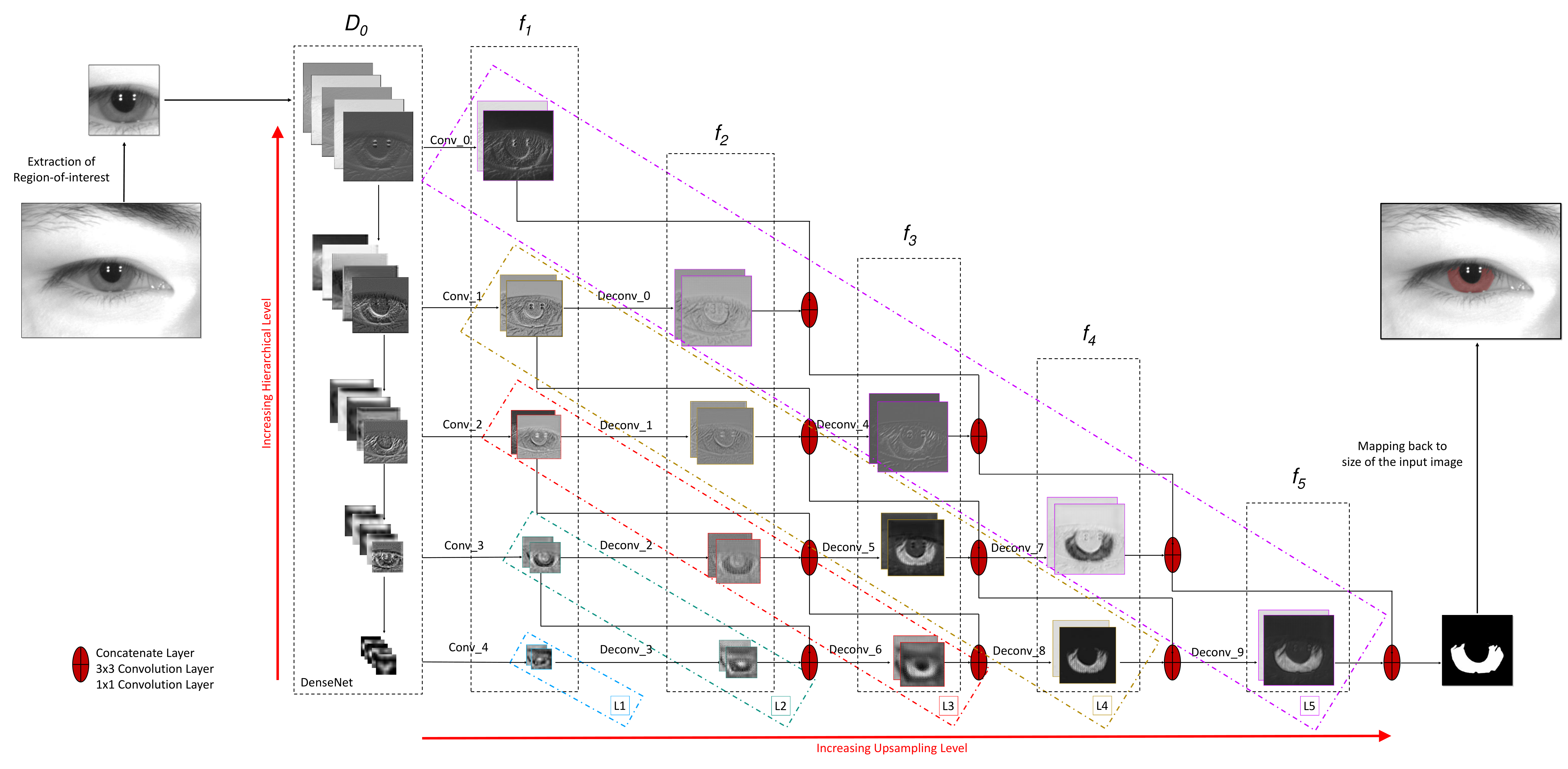}
\caption{Presents the proposed architecture, $PyramidNet$ for iris and pupil segmentation in an unconstrained environment. The dotted boxes represent the pyramid structure. The upsampling level increases in the x-direction, and the hierarchical level increases in the y-direction. The intermediate feature maps in L1-L5 levels present the different information stored in each map which results in preserving the fine and the global structure of the iris and pupil in the final output. (Best viewed in color)}
\label{fig:fig_arch}
\end{figure*}

\section{Proposed MTCD Approach}
%This section describes the iris segmentation and the classification approach used in the paper 

The broad pipeline of the proposed MTCD algorithm is shown in Fig. \ref{fig:pipeline}. The eye image acquired from the NIR camera is given as input to the segmentation network. The segmented image is then used by the multitask network for classification. In this pipeline, the segmentation algorithm has to be robust to address real-world challenges such as specular reflections, eyelashes, de-pigmentation, irregularities due to cataract and cataract-removal surgery. In this section, we present the proposed PyramidNet for iris and pupil boundaries segmentation followed by the classification network.

\begin{figure}[!t]
\centering
\includegraphics[scale = 0.4]{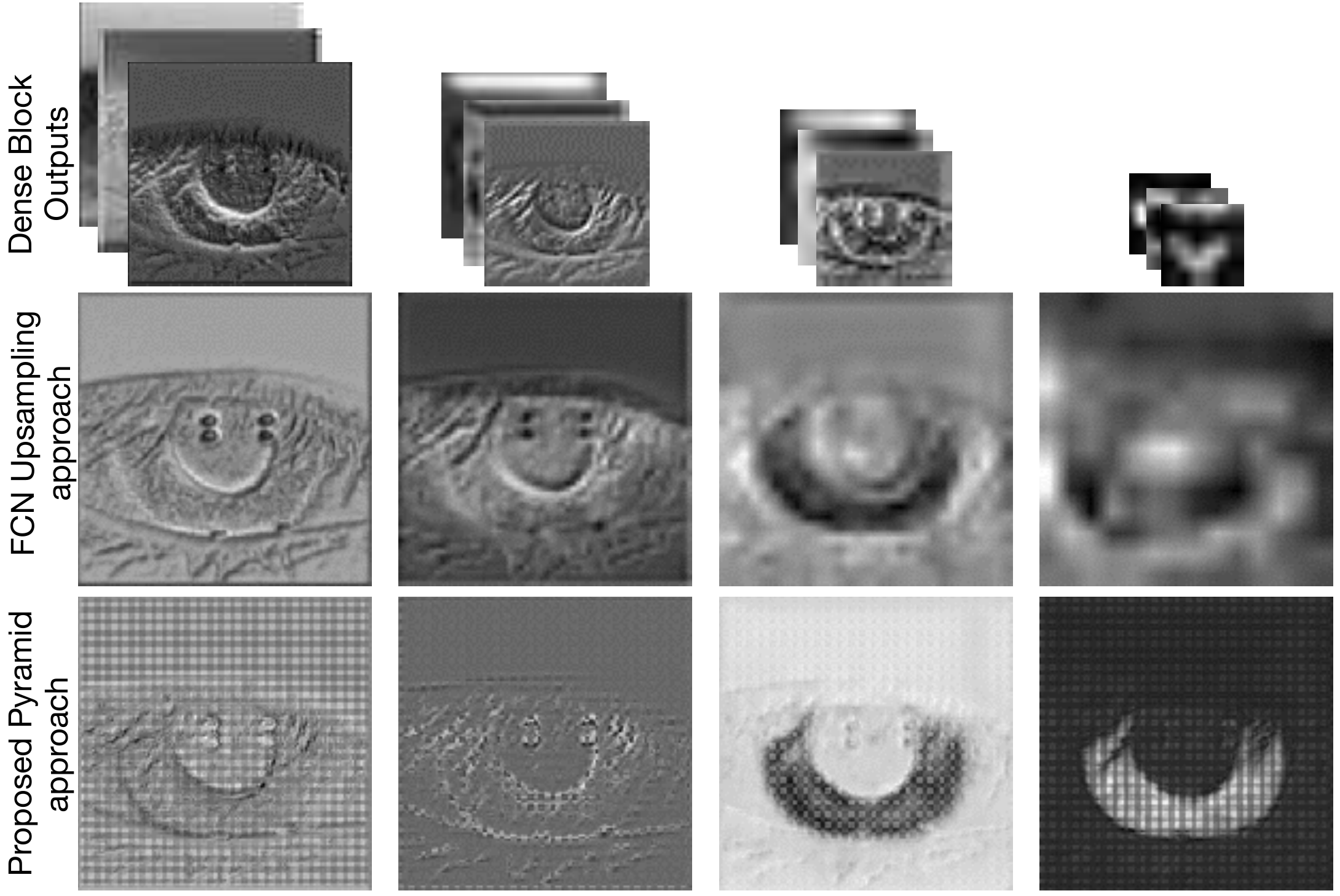}
\caption{Illustrating the difference in the local information present in the intermediate outputs when the feature maps are directly upsampled compared to the when the upsampling is done in the proposed pyramid like fashion.}
\label{inter_output}
\end{figure}

\subsection{Proposed PyramidNet for Iris and Pupil Segmentation}
%\subsection{Preliminaries}

Fig. \ref{fig:fig_arch} presents a diagrammatic representation of the proposed algorithm. The input image is processed by the proposed algorithm which produces its binary mask. This mask is multiplied with the original input image to extract the region of interest with iris and pupil boundaries. 

The proposed algorithm uses DenseNet as the backbone network. To improve the flow of information and gradient between the layers, \citep{huang2017densely} proposed dense connectivity between the layers (direct connections from any layer to subsequent layers). Let the input image be $I_{0}$ and $P_{t}(\cdot)$ be the non-linear transformation of the $t^{th}$ layer. Input to the $t^{th}$ layer is a concatenation of all the feature maps from the preceding layers, $I_{0}, I_{1}, ..., I_{t-1}$, i.e., %Mathematically it can be represented as: 
\begin{equation}
\centering I_{t} = P_{t}(c(I_{0}, I_{1}, ..., I_{t-1}))
\end{equation}
where, $c(I_{0}, I_{1}, ..., I_{t-1})$ denotes the concatenation of the of the feature-maps produced in layers $0,...,(t_{t-1} )$. To allow down-sampling of the feature maps, the DenseNet architecture has been divided into multiple densely connected blocks known as \textit{dense blocks}. We represent the set of these \textit{dense blocks} as $D_{0}^{i}$ where the range of $i$ is from $1$ to the total number of \textit{dense blocks}. For the task of image classification, DenseNet is trained using categorical cross-entropy loss function. In the proposed method, DenseNet has been used in the \textit{Pyramid Structure} for iris and pupil segmentation.

%\subsection{}

\vspace{15pt}
\noindent\textbf{Upsampling using Pyramid Structure:} Deep learning architectures \citep{arsalan2017deep, arsalan2018irisdensenet, lakra2018segdensenet, liu_MFCN, FCN}, directly upsample the intermediate outputs to the size of the final predicted mask resulting in a coarse mask. Upsampling one resolution up, fusing with the previous intermediate output, and continuing the upsampling process in this manner preserves the finest details. For instance, if the feature map of size $n\times n$ is directly upsampled to $4n \times 4n$, then the local structure is not fully preserved. However, if the $n\times n$ feature map is first upsampled to $2n \times 2n$ followed by an upsampling to the size, $4n \times 4n$, then the maximum local structure is preserved. We refer to this kind of upsampling procedure as upsampling in a pyramidic manner. Fig. \ref{inter_output} presents the difference in the feature maps fused to create the final output.

As shown in Fig. \ref{fig:fig_arch} we first reduce the number of channels of each dense block, $D_{0}^{i}$ to two and consider the segmentation as a two-class semantic segmentation problem, viz. iris class and background class. This creates the first \textit{deep pyramid} structure. Each deep pyramid is represented as $f^{i}_j$ where $i$ denotes the hierarchy of the feature maps in the y-direction, and $j$ represents the upsampling level in the x-direction. Stacking multiple such \textit{deep pyramids} creates a \textit{structural pyramid} where each level is represented as \textit{Lr}, where $r$ is equal to the number of \textit{deep pyramids}.

\textbf{\textit{Deep Pyramid:}} As represented in Equation \ref{eq:1}, the output of each of the dense blocks is convolved with a $1\times1$ kernel to reduce the number of channels to two. %The set of feature maps containing the intermediate outputs of each dense block is represented as $[D^{1}_0,\cdots, D^{nBlocks}_0]$ or $D^i_0$. 
This convolution operation sets the beginning point of our upsampling path and creates the first \textit{deep pyramid}, symbolized as $f^i_1$ where the range of \textit{i} is from $1$ to the number of outputs in a hierarchical level, in this case, the maximum value of \textit{i} is equal to the number of blocks present in the base architecture. Mathematically, $f^i_1$ represents the set of feature maps present in this level, $[f^{1}_1,\cdots,f^{nBlocks}_1]$, where \textit{nBlocks} is equivalent to the number of dense blocks in the base architecture.

\begin{equation}
f{_{1}}^{i} = Conv_{1\times1}(D{_{0}}^{i}), \;\\ where, i\in [1,\cdots,nBlocks]
\label{eq:1}
\end{equation}

The next \textit{deep pyramid}, whose set of feature maps is represented as $[f^{1}_2,\cdots,f^{nBlocks-(j-1)}_2]$ utilizes $[f^{1}_1,\cdots,f^{nBlocks}_1]$. It is mathematically defined as:

\begin{multline}
f{_{2}}^{i} = f{_{1}}^{i} \odot Deconv(f{_{1}}^{i+1}), \; \\ where, i\in [1,\cdots,nBlocks-1]
\label{eq:5}
\end{multline}

\noindent where the $\odot$ symbol denotes a set of fusion operations to combine the feature maps. After deconvolution, the upsampled features maps are concatenated with the features maps of the previous hierarchical level. After this, a $3\times3$ convolution filter is applied to this two-channel output. This convolution operation is done for two reasons. Firstly, it reduces the aliasing effect that may have occurred due to upsampling of lower resolution feature maps. Secondly, it helps in removing the noise present in the higher resolution feature maps. Due to the concatenation operation, the number of channels in the fused output increases from two to four. To reduce the number of channels back to two, we apply $1\times1$ convolution on the fused output. We continue fusing the outputs of each of the \textit{deep pyramid} until the hierarchy level becomes the same as the number of blocks. Mathematically, every \textit{deep pyramid} can be defined as:
\\
\begin{multline}
f{_{j}}^{i} = f{_{j-1}}^{i} \odot Deconv(f{_{j-1}}^{i+1}) where, \\ j \in [2,\cdots,nBlocks], i\in [1,\cdots, nBlocks-(j-1)] 
\label{eq:6}
\end{multline}

\noindent where $i$ denotes the hierarchy of the feature maps in the y-direction, and $j$ represents the upsampling level in the horizontal upsampling path. Due to the fusion of feature maps, the number of hierarchy levels keeps decreasing as we move forward in the horizontal upsampling path. 

\begin{figure}[!t]
\centering
\includegraphics[width=3.35in]{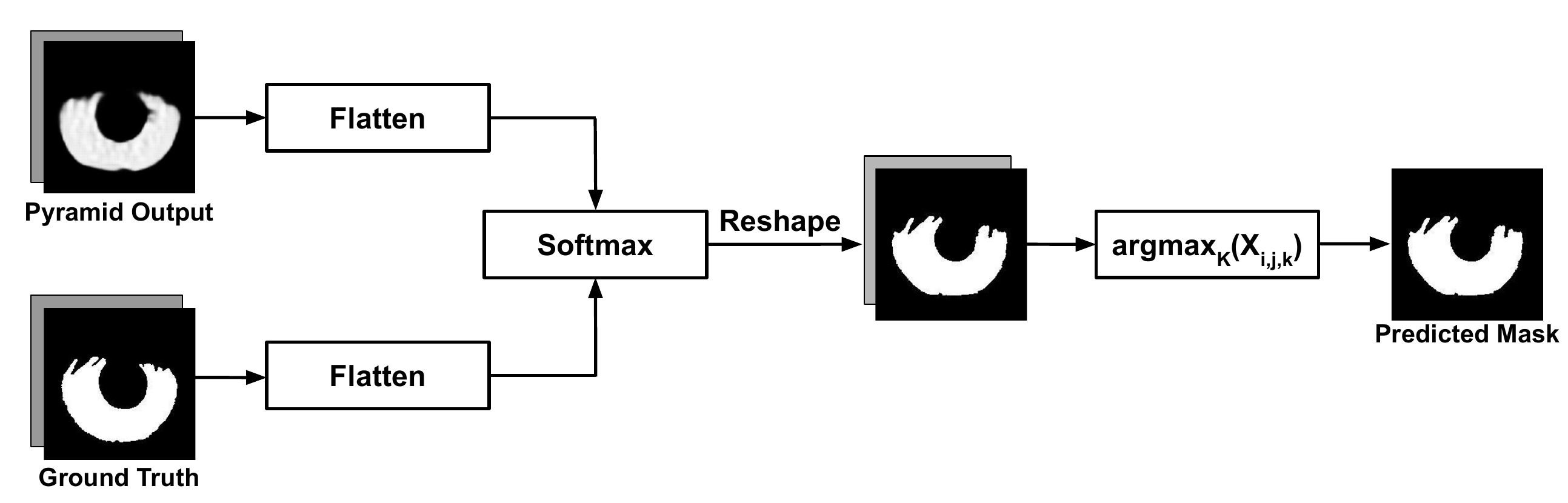}
\caption{Illustrating the process of calculating the loss that is back-propagated through the segmentation network.}
\label{loss_fnctn}
\end{figure}

As shown in Fig. \ref{fig:fig_arch} it can be observed that each \textit{deep pyramid} contains varied information. The feature map set of the highest hierarchical level has the maximum resolution. It contains maximum noise along with very fine details of the iris. The last hierarchical level feature map set of least resolution contains minimum noise and preserves the maximum global iris and pupil structure. Hence, when these feature maps are fused to create the next \textit{deep pyramid}, the maximum amount of noise is removed while keeping the fine and global iris and pupil structure intact. Further, the total computation cost while adding the feature maps is minimal. Consequently, accurate masks can be obtained without introducing too many overhead parameters to the base network.

\textbf{\textit{Structural Pyramid:}} Fusing all the \textit{deep pyramids} in the proposed manner creates a \textit{structural pyramid}. Each level of \textit{structural pyramid} contains feature maps of the same resolution and is represented as \textit{Lr}, where r is equivalent to the number of \textit{deep pyramids}. It can be visually seen from Fig \ref{fig:fig_arch} that each set of feature maps in a particular \textit{structural pyramid} level presents different information towards the final prediction. For instance, in level $L5$ (represented in Magenta), some feature maps preserve the edge information, whereas others preserve the global structure of the iris and pupil, resulting in an accurate mask. 
%\vspace{15pt}

\noindent \textbf{Iris and Pupil Mask Prediction:} Once we have only one set of feature maps in the \textit{deep pyramid}, it is flattened and softmax is applied to obtain per-pixel classification, i.e. %as shown in equation \ref{eq:7}. 

\begin{equation}
P(y=j | \Theta^{(i)}) = \frac{exp(\Theta^{(i)}x_{(p,q)})}{\sum_{j=0}^{k}exp(\Theta_{k}^{(i)}x_{(p,q)})}
\label{eq:7}
\end{equation}

\noindent where, $k$ represents the number of classes, viz. two in our case and $\Theta^{(i)}$ symbolizes the softmax parameters and the probability map, $P(y=j | \Theta^{(i)})$ is achieved. To get the eye mask, each pixel is allocated the channel with the highest probability. Fig. \ref{loss_fnctn} illustrating the process of predicting the mask. Finally, a binary morphological post-processing is performed where the mask is first dilated, followed by erosion operation. Finally, the eroded output is multiplied with the original image to generate a region of interest, i.e. eye region only.

%\vspace{15pt}

\subsection{Cataract Classification using Multitask Learning}

\begin{figure}[!t]
\centering
\includegraphics[width=3.35in]{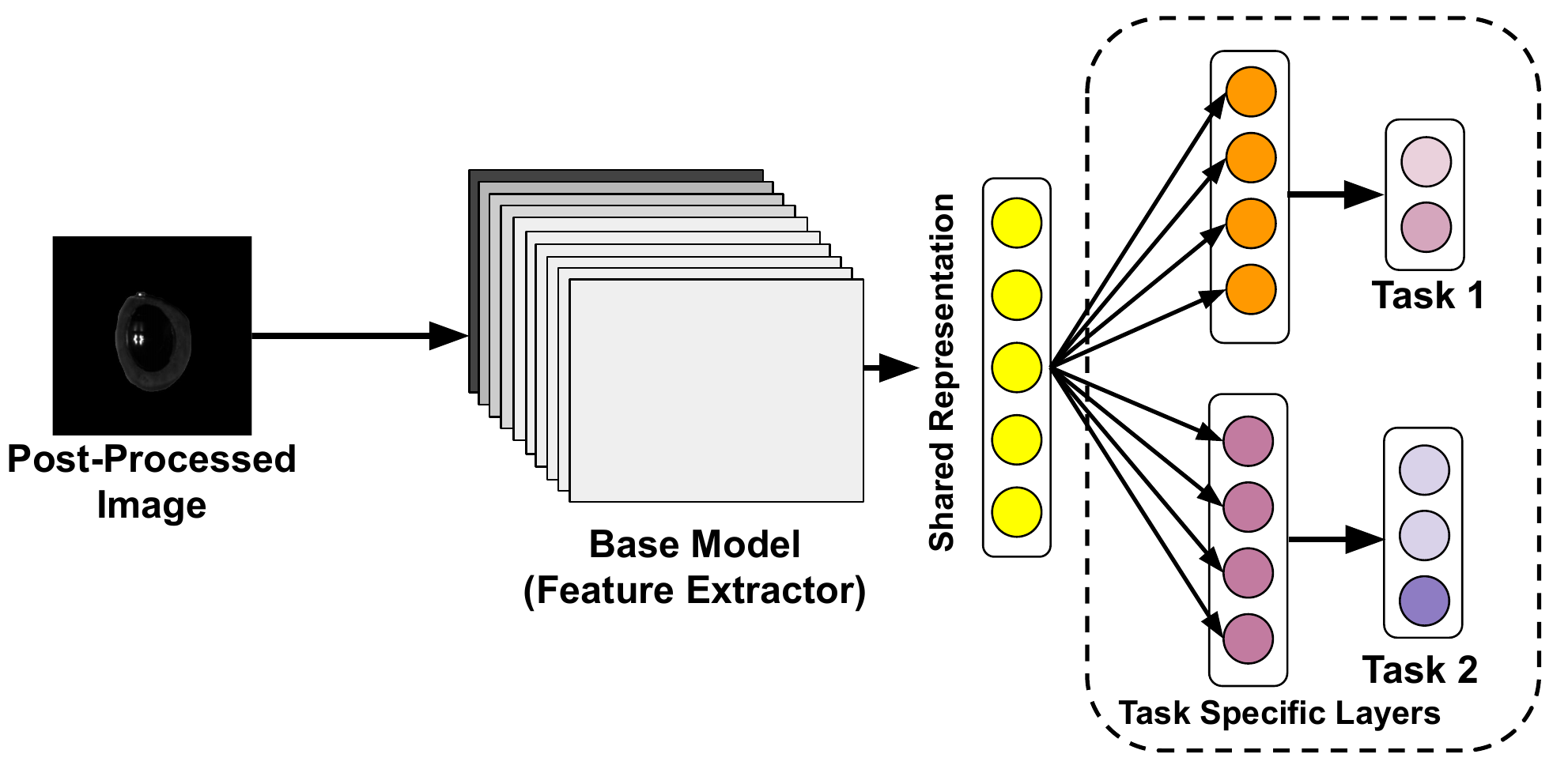}
\caption{Multitask Classification Network for Cataract Classification.}
\label{fig:Class}
\end{figure}

For the given problem of cataract detection, a segmented eye can be healthy or unhealthy and if unhealthy, it can be a cataract or any other disorder. The cataract affected eye may further be categorized into pre-cataract surgery or post-cataract surgery. In this research, we present this problem as a multitask learning problem with the following two tasks:    

\begin{itemize}
    \item Task 1 (T1): the first task is to classify the input image into one of the classes: $y_{T_1}\in$ \{\textit{healthy}, \textit{unhealthy}\}
    \item Task 2 (T2): second task is to classify the input image into three classes, i.e. $y_{T_2}\in$ \{\textit{pre-cataract}, \textit{post-cataract}, \textit{others\footnote{The 'others' class consists of samples which are neither affected by cataract nor by surgery.}}\}. 
\end{itemize}

Multitask learning can be accomplished in various ways, such as joint learning of multiple related tasks \cite{liu2019joint} and learning auxiliary tasks to support main task \citep{liebel2018auxiliary}. Due to a limited number of images in the dataset, it is not easy to train any deep neural network from scratch. Therefore, we have used the transfer learning approach resulting in a reduction of training (computation) time as well as help in achieving better performance on smaller datasets. We have used the pre-trained ResNet50 \citep{he2016deep} as the base model and trained it for learning feature representations for cataract detection. 

%\textbf{Loss function:} 
Fig. \ref{fig:Class} illustrating the block diagram of the proposed multitask network. To train this network, joint optimization of the losses pertaining to these two tasks are performed. The final loss function is computed as the weighted sum of two classification losses. We have used binary cross-entropy (BCE) loss and categorical cross-entropy (CCE) loss for Task 1 and Task 2, respectively. The two individual losses and the final loss are defined as follows: 

\begin{equation}
    BCE = -y_{T_1}log(p)-(1-y_{T_1})log(1-p)
\end{equation}
\begin{equation}
    CCE = -\sum_{i=1}^{3}y_{T_2}^{i}log(y_{T_2}^{i})
\end{equation}
\begin{equation}
    Final loss = \lambda*BCE + CCE
\end{equation}
where, \textit{i} is the class index and \textit{p} is the class probability.  

\subsection{Implementation Details}

This section provides the implementation details of the proposed approach. 

\textbf{Segmentation Network:} The proposed segmentation architecture, $PyramidNet$ utilizes the DenseNet model with $43$ convolution layers and is trained from scratch using the CASIAv4-distance dataset \footnote{http://biometrics.idealtest.org/dbDetailForUser.do?id=4}, UBIRISv2 \citep{ubiris} and IIITD Cataract Surgery dataset \citep{nigamphacoemulsification}. The model is trained for 60 epochs using adaptive moment estimation \citep{kingma2014adam}, \textit{Adam} optimiser with initial learning rate of $0.001$. 
 
During training, contrast normalization and flip operations are used to augment the dataset size by $10$ times. For contrast normalization, $5$ different contrast factors have been used. %The number of times by which the difference between a pixel value and the center value is multiplied is called as the contrast factor. 
Size of the input images for all the datasets in the NIR spectrum is $640\times480$. The ROI is extracted using SegDenseNet \citep{lakra2018segdensenet}. After extraction of ROI the size of the image reduces to $224 \times 224$ which is then fed into the proposed $PyramidNet$. 

\textbf{Classification Network:} For training the classification network, we have used IIITD Cataract Surgery, IIITD alcohol and (the proposed) pupil dilation datasets. The cataract samples (pre-cataract and post-cataract surgery) are considered as unhealthy for Task 1 and then two separate classes in Task 2. The other two datasets are used as the healthy class  (more details about the dataset are in the next section).     
For Task 1 and Task 2, we have used sigmoid and softmax activation functions, respectively. 
For feature extraction, transfer learning concept is utilized where pre-trained (on ImageNet dataset) ResNet50 is used as the base model and fine-tuning is performed on the train sets of the above mentioned datasets. As shown in Fig. \ref{fig:fig_arch}, a global average pooling (GAP) layer and two fully connected (FC) layers are added on the pre-trained ResNet50 model. These two fully connected layers are added for two classification tasks, Task 1 and Task 2. The best results are obtained with a model trained on 100 epochs with a learning rate of 0.00001, $\lambda = 0.5$, Adam as an optimizer, and a batch size of 4 on NVIDIA V100 32GB GPU. To achieve better generalization, we have also performed data augmentation with contrast normalization by various factors and flip operations, which increased the dataset size by five times. 

 \section{Datasets}
 
The proposed deep learning based segmentation and classification method is evaluated on three datasets, viz., IIITD Cataract Surgery \citep{nigamphacoemulsification}, IIITD Alcohol \citep{arora2012iris}, and on the proposed Pupil Dilation dataset. These datasets are chosen since they comprise various covariates of eye image, making them suitable choices for evaluating the efficiency of the proposed models.

\vspace{6pt}

\begin{figure}[!t]
\centering
\includegraphics[width=0.48\textwidth]{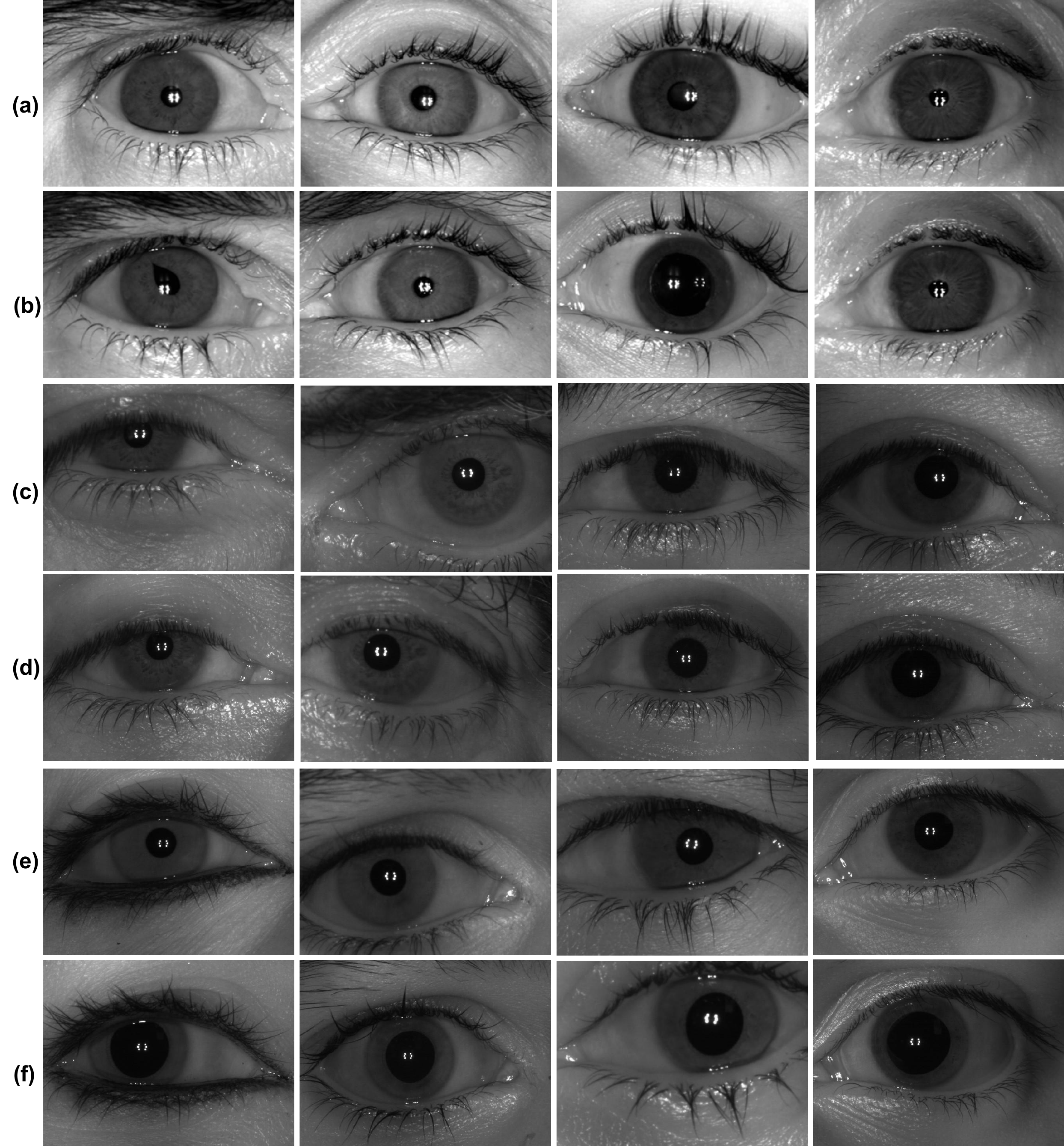}
\caption{Sample images: (a) and (b) are pre and post cataract surgery; (c) and (d) are pre and post alcohol; (e) and (f) are pre and post pupil dilation from the Pupil Dilation dataset.}
\label{fig:fig_samples_pupil_dilation}
\end{figure}

\noindent \textbf{Pupil Dilation dataset:} The proposed dataset contains images showcasing variations due to Pupil Dilation. Tropicacyl Plus, a prescription drug used to treat paralysis of the ciliary muscle and dilate pupils before and after ophthalmic surgery, is used to create the dataset. The dataset consists of $528$ images acquired from human subjects before and after the medicine is administered by the ophthalmologist. The pupil dilation dataset contains 528 images, $264$ pre-eyedrop treatment and $264$ post-eyedrop treatment images of $44$ subjects. Fig. \ref{fig:fig_samples_pupil_dilation} shows sample images acquired pre and post eyedrop treatment. Table \ref{pupil_dilation} presents various characteristics of the images. To the best of our knowledge, this is the first dataset of it's kind and it will be releasing it to the research community. For experiments, 50 samples are used for testing while the remaining form the training set. After augmentation, the number of training samples is 1815.

\vspace{6pt}
\noindent \textbf{IIITD Cataract Surgery Dataset} contains 880 samples from 132 individuals, 440 each representing cataract and post cataract surgery samples (represented as pre and post cataract surgery). 100 samples from both the classes are kept in test set and the remaining comprise the train set. After augmentation, the number of training samples is 4080.

\vspace{6pt}
\noindent \textbf{IIITD Alcohol dataset}  \cite{arora2012iris} studied the effect of alcohol on pupil dilation/constricts. The pupil dilates/constricts due to intake of alcohol which results in affecting the iris recognition performance. Also, it is clearly shown in Fig. \ref{fig:fig_samples_pupil_dilation} (row c and d) how the alcohol can affect the size of the pupil which in turn can affect the iris recognition. More details about this dataset can be found in \cite{arora2012iris}. This dataset contains 440 images pertaining to 110 subjects. Of these, 50 randomly selected samples are used for testing, while the remaining comprise the training set. After applying the augmentation, the number of training samples is 1170.

\vspace{6pt}
\noindent \textbf{Data Preparation:} For learning the segmentation model, we have pre-trained the model on CASIAv4-distance and UBIRISv2 \citep{ubiris} datasets and then IIITD Alcohol and IIITD Cataract Surgery datasets are used for fine-tuning. For learning the cataract classification model, the ImageNet pre-trained base model is used and then IIITD Alcohol, IIITD Cataract Surgery, and the proposed Pupil Dilation datasets are used. Further, data augmentation is applied so as to minimize the data imbalance problem. IIITD Alcohol and Pupil Dilation datasets belong to one class, and the IIITD Cataract Surgery belongs to the other class, thus making the overall data balanced. Ground truth segmentation masks for iris and pupil have been manually annotated using Adobe Photoshop. We will release the proposed database, annotations, and train-test partition details via \url{http://iab-rubric.org/}.

\begin{table}[!t]
% \scriptsize
\centering
\caption{Characteristics of the proposed Pupil Dilation dataset.}
\label{pupil_dilation}
\begin{tabular}{|l|l|}
\hline
\textbf{Characteristics} & \textbf{Pupil Dilation}                                                                                  \\ \hline
Sensor                   & Vista Sensor                                                                                                             \\ \hline
Environment              & Indoor                                                                                                             \\ \hline
Sessions                 & Two                                                                                                          \\ \hline
%Attributes of subjects   &                                                                                                              \\ \hline
No. of individuals  & 88                                                                                                           \\ \hline
No. of images            & 276 (pre) and 276 (post)                                                                                          \\ \hline
Resolution               & 640x480                                                                                                      \\ \hline
Challenges               & \begin{tabular}[c]{@{}p{4cm}@{}}Excessive dilation due to the administered eyedrops.\end{tabular} \\ \hline
\end{tabular}
\end{table}

\begin{comment}

\subsection{Existing datasets}
%\textcolor{red}{YA, MK: Add the statistics of other datasets as well.} 
Table \ref{fig:exp_prot} presents the number of samples in each dataset used in this work. Except for the UBIRISv2 dataset, which is present in the visible spectrum,  we used the NIR images from the other datasets in this work.

\end{comment}

\section{Experimental Results}
The performance of the proposed MCTD approach is presented in two parts, (i) segmentation and (ii) classification. The effectiveness of the algorithm is compared by varying the base model and comparing the results with existing algorithms. We have also performed an ablation study to demonstrate the effectiveness of various components of the algorithm.

\subsection{Segmentation Performance}

The performance of the segmentation algorithm is measured using the average classification error rate proposed in the NICE-I competition \citep{NICE}. 
\begin{equation}
Error = \frac{1}{N\times m\times n} \sum_{i,j = 1}^{m,n} M_G^{te'}(i,j)\oplus M_P^{te'}(i,j)
\label{eq:metric}
\end{equation}

\noindent where, $M_G^{te'}$, $M_P^{te'}$, $N$, $m$ and $n$ denote the ground truth mask, the predicted mask, total number of test samples, height, and width of the mask, respectively. The logical exclusive-OR operator calculates the correspondent disagreeing pixels' proportion between the ground truth and the predicted segmentation mask. We compare results with a non-deep learning method~\citep{ICCV_2015} and two deep learning methods: IrisParseNet \citep{wang2020towards}, and SegDenseNet~\citep{lakra2018segdensenet}.

\begin{figure}
\centering
\includegraphics[width=0.4\textwidth,height=4cm,keepaspectratio]{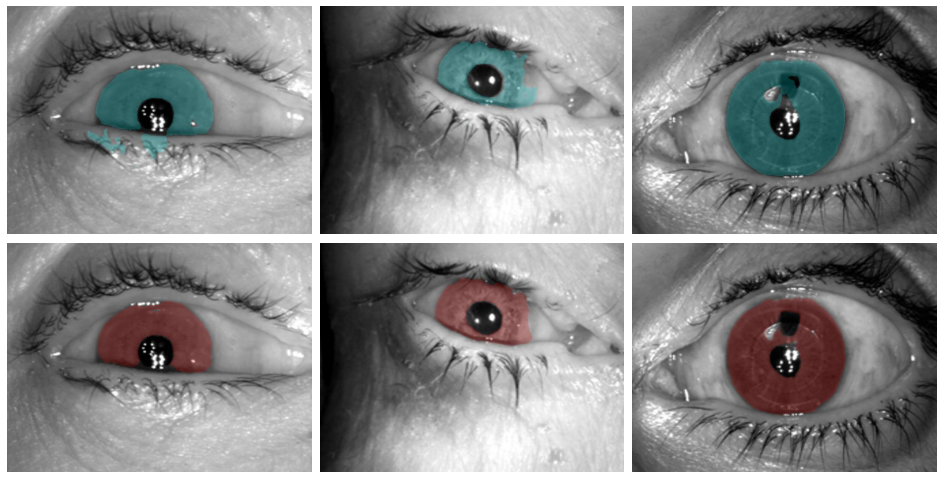}
\caption{Illustrating the segmentation output by FCN-8s (first row) and the proposed $PyramidNet$ (second row) algorithms on the IIITD Cataract Surgery dataset. The masks are overlaid on the images to visually demonstrate segmentation with respect to the iris and pupil boundaries. The results demonstrate that the proposed algorithm yields finer boundaries compared to FCN-8s approach.}
\label{fig:fig_proposed_visual}
\end{figure}

\begin{figure}
\centering
\includegraphics[width=0.48\textwidth,height=4cm,keepaspectratio]{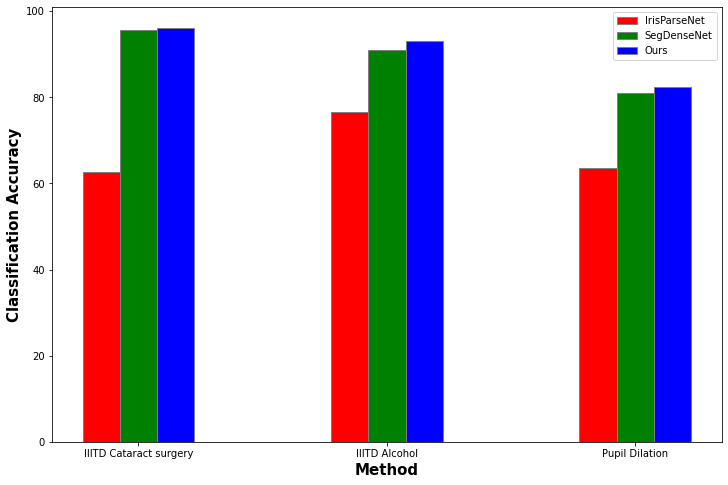}
\caption{Showcasing the classification accuracy of existing and proposed segmentation methods on the datasets used in the paper.}
\label{fig:cviu_barplot}
\end{figure}

Fig. \ref{fig:fig_proposed_visual} shows the sample results on the IIITD Cataract Surgery dataset where the masks are overlaid on the iris and pupil region. These examples show that the proposed algorithm is able to detect the fine boundaries of iris and pupil region. Table \ref{segmentation-accuracy-table2} presents segmentation errors obtained from the proposed algorithm and the existing algorithms. The percentage error has reduced by $21.4$\%, $11.9$\%, and $31.8$\% (from the next best performing model on these datasets) on IIITD Cataract Surgery, IIITD Alcohol, and Pupil Dilation datasets, respectively compared to existing techniques. It shows that our method yields state-of-the-art accuracies on all these datasets. Further, Fig. \ref{fig:cviu_barplot} compares the performance across the methods based on the classification accuracy. It can be observed that the proposed method is able to classify each iris pixel more accurately compared to existing deep learning methods for iris segmentation.

\begin{table}[]
%\scriptsize
\centering
\caption{Comparisons of the proposed and existing iris segmentation techniques using average segmentation error (\%). For fair comparison no post-processing is done for \cite{wang2020towards}.}
\label{segmentation-accuracy-table2}
\begin{tabular}{|l|c|c|c|}
\hline
\textbf{Method}                                                      & \multicolumn{1}{l|}{\textbf{\begin{tabular}[c]{@{}l@{}}IIITD \\ Cataract \\ Surgery \end{tabular}}} & \textbf{\begin{tabular}[c]{@{}l@{}}IIITD \\ Alcohol \end{tabular}} & \textbf{\begin{tabular}[c]{@{}l@{}} Pupil \\ Dilation\end{tabular}}  \\ \hline
\begin{tabular}[c]{@{}l@{}}IrisParseNet, 2020\\ \citep{wang2020towards}\end{tabular}  & 9.87   & 3.06    & 8.16   \\ \hline

\begin{tabular}[c]{@{}l@{}}Zhao and Kumar, 2015\\ \citep{ICCV_2015}\end{tabular}                                                                 & 6.28 & 8.51   & 7.67   \\ \hline
\begin{tabular}[c]{@{}l@{}}SegDenseNet, 2018\\ \citep{lakra2018segdensenet}\end{tabular} & 0.98   & 1.42    & 3.46    \\ \hline
% \begin{tabular}[c]{@{}l@{}}Liu et al. (HCNNs), 2016\cite{liu_MFCN}\end{tabular}            & 1.08   & -    & -   & -   & 1.11              \\ \hline
% \begin{tabular}[c]{@{}l@{}}Liu et al. (MFCNs), 2016\cite{liu_MFCN}\end{tabular}            & 0.59      & -      & -      & -    & 0.90              \\ \hline
\textbf{\begin{tabular}[c]{@{}l@{}}Proposed Method\end{tabular}}                                                             & \textbf{0.77}                                                                                     & \textbf{1.25}                                                     & \textbf{2.36}                                                                            \\ \hline
\end{tabular}
\end{table}

\begin{figure}
\centering
\includegraphics[scale=0.475]{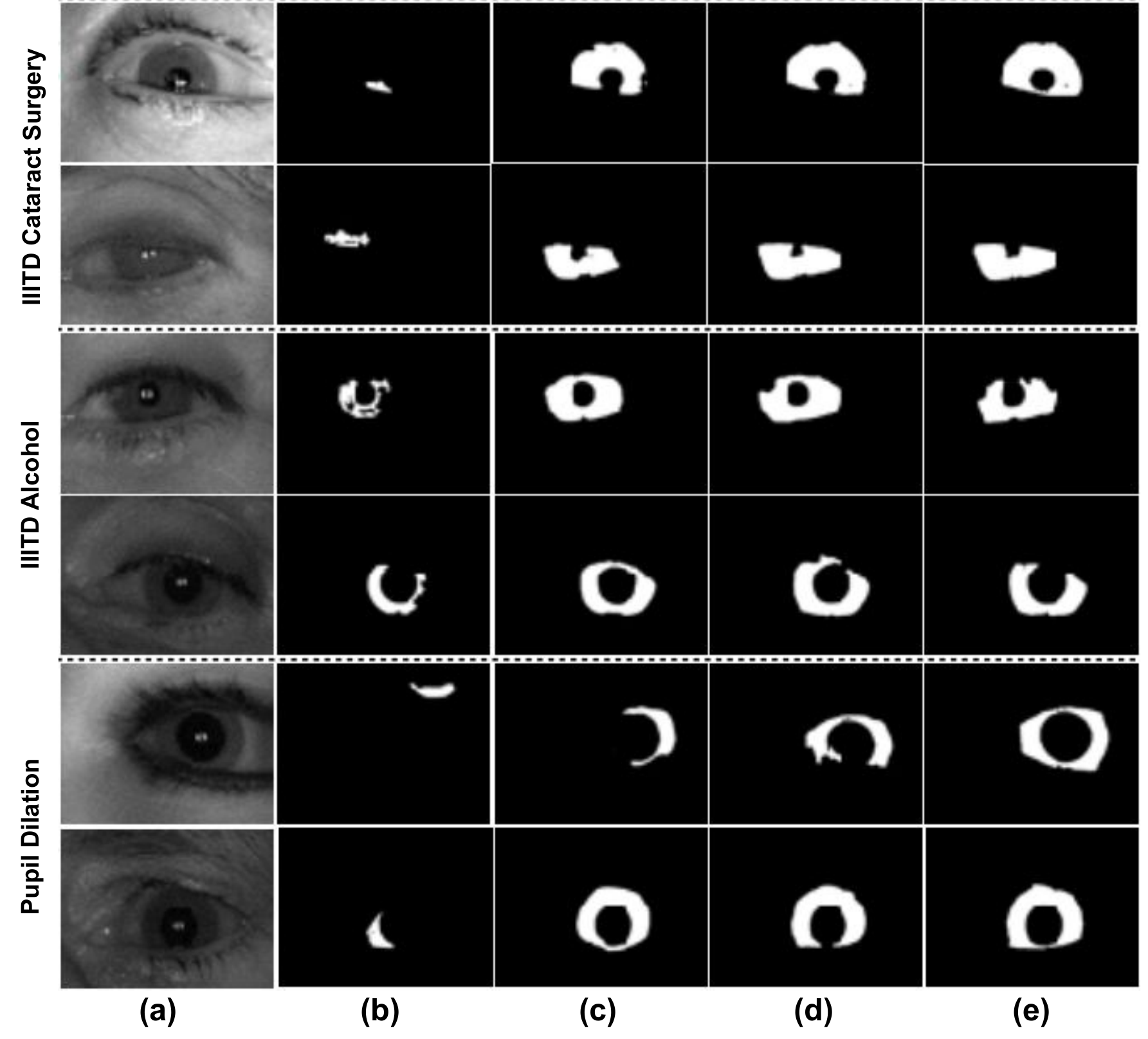}
\caption{Showcasing the results of iris segmentation on multiple datasets. (a) The input image; masks obtained by (b) \cite{ICCV_2015} method, (c) SegDenseNet \citep{lakra2018segdensenet} (the next best performing deep learning approach), (d) proposed PyramidNet, and (e) ground truth.}
\label{fig:fig_results}
\end{figure}

 Fig. \ref{fig:fig_results} shows sample masks generated by the proposed method and comparison with existing algorithms on the three datasets. As can be visually observed, the proposed method can predict very accurate masks, implying that the proposed method preserves both global and fine structures of the iris and pupil. The first row shows how the model can segment the iris region even when it is severely occluded by reflection. Further, all the masks predicted by the proposed method have fine-details, such as removing areas secluded by fine eyelashes. Also, unlike the SegDenseNet \citep{lakra2018segdensenet}, the proposed method can predict the mask for sample images of IIITD Cataract dataset, which contains bubbles. It is our assertion that the proposed method can overcome this because upsampling in pyramid fashion preserves both the local and global structure. Further, as shown in Fig. \ref{fig:fig_results}, when the contrast difference between the iris and the sclera region is extremely low, the proposed algorithm is still be able to detect the boundaries. It can be directly observed that both SegDenseNet \citep{lakra2018segdensenet} and \citep{ICCV_2015} fail to segment the boundaries correctly. However, the proposed method, $PyramidNet$ can handle these cases with great precision because it restores the information in a pyramid manner. The fine edge information and global structure present in the \textit{structural pyramid} $L5$ when fused can accurately predict the boundaries even when the contrast difference is extremely low.

We also compare the performance of the proposed algorithm with the FCN architecture~\cite{FCN}. In this approach, a deconvolution operation has been used to upscale the image and combine with the previous layer output feature maps. However, in the proposed $PyramidNet$ architecture, the second and third blocks of the DenseNet (as shown in figure~\ref{fig:fig_arch}) are utilized in two ways. This results in multiple feature maps of the same resolution. Combining these incorporates both the coarse and fine structures of iris and pupil in the segmentation process. The difference between the proposed $PyramidNet$ and FCN-8s outputs has been shown in Figure~\ref{fig:fig_proposed_visual}. The results for FCN-8s are computed using our implementation of FCN-8s. On the cataract dataset, FCN yields $1.35$\% segmentation error and $PyramidNet$ achieves $0.77$\% segmentation error. This comparison shows that for iris and pupil boundary segmentation, it is imperative to combine feature maps at each upsampling level.

To further show the efficacy of each component of $PyramidNet$ architecture, ablation study is performed. In the proposed method the number of \textit{structural} levels is equivalent to the number of dense blocks present in the base architecture. To understand the effect of each \textit{structural} pyramid level on the final output, we have computed segmentation error. In our ablative study, the least segmentation error is achieved when all the dense blocks are used for building the \textit{structural} pyramid. This is so because for iris and pupil segmentation both global and fine structures must be preserved. It is our assertion that maximum fine structure is preserved by the output of the first level of the \textit{structural} pyramid and the maximum global information is stored in the last i.e. fifth level of the \textit{structural} pyramid. We have observed  that there is a small decrease in segmentation error if the feature maps obtained at \textit{structural} pyramid level $L3$ are directly upsampled to the size of the output image compared to upsampling of feature maps of $L2$ level. However, on upsampling the feature map obtained at $L5$ level, there is a significant decrease in the segmentation error because the maximum amount of local information/the finest details of the iris and pupil are preserved in it.

\begin{comment}

\begin{table}[!t]
\centering
\caption{Average segmentation error (\%) with different number of pyramid levels.}
\label{pyramid_level}
\begin{tabular}{|p{2.5cm}|c|}
\hline
\textbf{Method}                                                      & \textbf{\begin{tabular}[c]{@{}c@{}}CASIAv4 distance\end{tabular}} \\ \hline
L2 level                                                                         & 0.47                                                                \\ \hline
L3 level                                                                        & 0.46                                                                \\ \hline
L4 level                                                                         & 0.44                                                                \\ \hline
\begin{tabular}[c]{@{}c@{}}\textbf{L5 level} \end{tabular}                   & \textbf{0.40}                                                        \\ \hline
\end{tabular}
\end{table}

\end{comment}

\begin{table}[!t]
\centering
\caption{Summarizes the performance of the proposed approach of multitask eye image classification by changing the segmentation algorithm. We used SegDenseNet \citep{lakra2018segdensenet} as a baseline and replaced it with our proposed PyramidNet, which outperformed the baseline results. It is also evident from the table that the post-processing step on the segmentation masks of PyramidNet improves both the tasks' overall performance.}
\label{tab:tableperf}
\resizebox{0.45\textwidth}{!}{
\begin{tabular}{|l|l|c|c|c|c|}
\hline
\textbf{\begin{tabular}[c]{@{}c@{}}Segmentation \\  Algorithms\end{tabular}}                                 &    & \textbf{\begin{tabular}[c]{@{}c@{}}Accuracy\\  (\%)\end{tabular}} & \textbf{Precision} & \textbf{Recall} & \textbf{\begin{tabular}[c]{@{}c@{}}F1 \\score\end{tabular}} \\ \hline
Baseline   & T1 & 97.34   & 0.96               & 0.97            & 0.97              \\ \cline{2-6} 
 SegDenseNet   & T2 & 92.34  & 0.92               & 0.92            & 0.92              \\ \hline
\multirow{2}{*}{{PyramidNet}}  & T1 & 100    & 1.0  & 1.0             & 1.0               \\ \cline{2-6} 
             & T2 & 95.67   & 0.96       & 0.96   & 0.96    \\ \hline
PyramidNet + & T1 & \textbf{100}  & \textbf{1.0}       & \textbf{1.0}    & \textbf{1.0}      \\ \cline{2-6} 
 Post-Processing   & T2 & {\ul \textbf{96.67}}                                              & \textbf{0.97}      & \textbf{0.97}   & \textbf{0.97}     \\ \hline
\end{tabular}}
\end{table}

\begin{table}[!t]
\centering
\caption{Characteristics of the models proposed for iris segmentation. Cost for \cite{wang2020towards} has been directly taken from the paper.}
\label{Compare_logistics}
\begin{tabular}{lccc}
\hline
\textbf{Algorithms}                                           & \textbf{\begin{tabular}[c]{@{}c@{}} Model \\size (MB)\end{tabular}} & \textbf{\begin{tabular}[c]{@{}c@{}}No. of\\ parameters (M) \end{tabular}} & \textbf{\begin{tabular}[c]{@{}c@{}}Test time \\(sec)\end{tabular}} \\ \hline
\begin{tabular}[c]{@{}l@{}}IrisParseNet\end{tabular} & 119.0      & 31.28      & 0.15  \\ \hline
\begin{tabular}[c]{@{}l@{}}SegDenseNet\end{tabular} & 57.30      & 8.00     & 0.024     \\ \hline
\textbf{Proposed}       & \textbf{11.9}   & \textbf{0.92}    & \textbf{0.017}                                                       \\ \hline
\end{tabular}
\end{table}

The proposed method has significantly lower number of parameters compared to IrisParseNet \citep{wang2020towards} and SegDenseNet \citep{lakra2018segdensenet}\footnote{Parameters are calculated using our own implementation of \citep{liu_MFCN} methods}. As shown in Table \ref{Compare_logistics}, the number of parameters has reduced by \textit{30 times} and the size of the model has reduced by \textit{10 times}. Further, the testing time of the proposed method is the least. The proposed model yields state-of-the-art results on three datasets and is the optimal model both in terms of computation cost and memory consumption. To be uniform, all the algorithms are implemented and run on the same machine, keeping all the configurations same.

\subsection{Cataract Classification}

The cataract classification performance is reported in terms of the classification accuracy, precision, recall, and F1 score. The output of segmentation algorithm, i.e. segmented iris and pupil region, is used as input to the classification algorithm. For comparison, we have used SegDenseNet \citep{lakra2018segdensenet} approach (2nd best segmentation approach - from Table \ref{segmentation-accuracy-table2}). Further, in order to showcase the effect of binary morphological operations (post-processing) after the proposed \textit{PyramidNet}, we have shown the results with and without post processing. Table \ref{tab:tableperf} summarizes the results of the proposed multitask classification algorithm with three segmentation approaches. It can be clearly observed that PyramidNet yields improved performance compared to the baseline results of SegDenseNet. PyramidNet differentiates between the healthy and unhealthy classes with 100\% accuracy. Further, PyramidNet with post-processing does not deteriorate the performance in Task T1 but improves the classification performance in Task T2. For differentiating between the diseased classes, i.e., task T2, PyramidNet with post-processing yields an error of only 3.3\%. Analyzing the precision and recall, we have observed that both precision and recall of \textit{healthy class} is 1. This result is due to the fact that there is no overlap between the samples of healthy and unhealthy classes. It is further supported by Table \ref{tab:CM} (confusion matrix) and 
%%and Fig. \ref{fig:tsne_plot} (TSNE plot). 
very few pre-cataract and post-cataract samples are misclassified into each other. Interestingly, among the remaining two classes, the precision of \textit{post-cataract class} is lower than the \textit{others} class, while the recall of the \textit{post-cataract class} is higher than the \textit{others} class. After post-processing, the overall performance and precision of post-cataract performance improves, however, the recall reduces marginally by 0.03.

\begin{figure}[!t]
\centering
\includegraphics[width=0.47\textwidth]{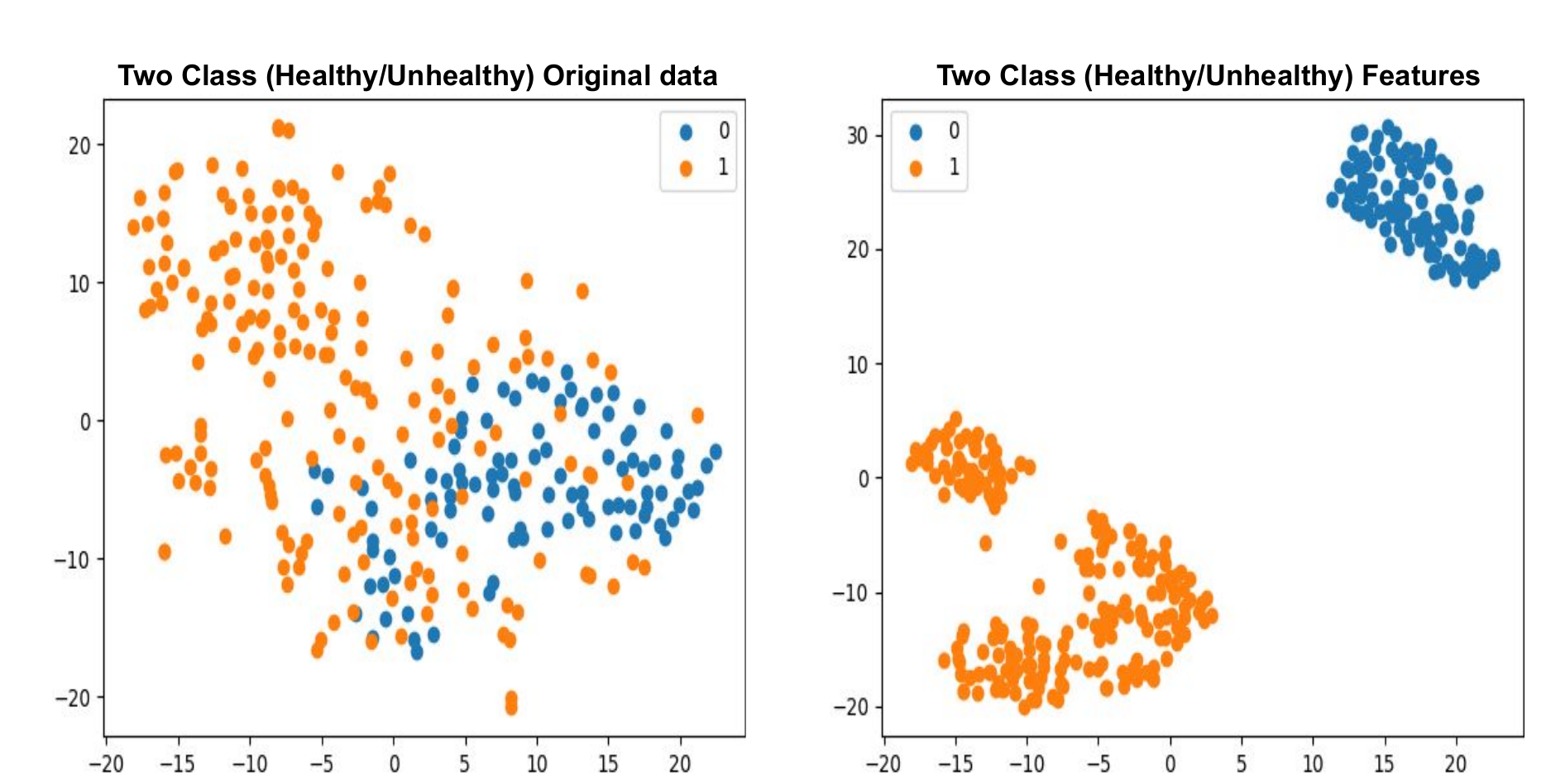}                
\caption{Illustrating the tSNE plot for Task 1: left plot shows the samples in original image spacea and the right plot shows the samples in feature space.}
\label{fig:tsne_plot}
\end{figure}

\begin{figure}[!t]
\centering
\includegraphics[width=3.4in]{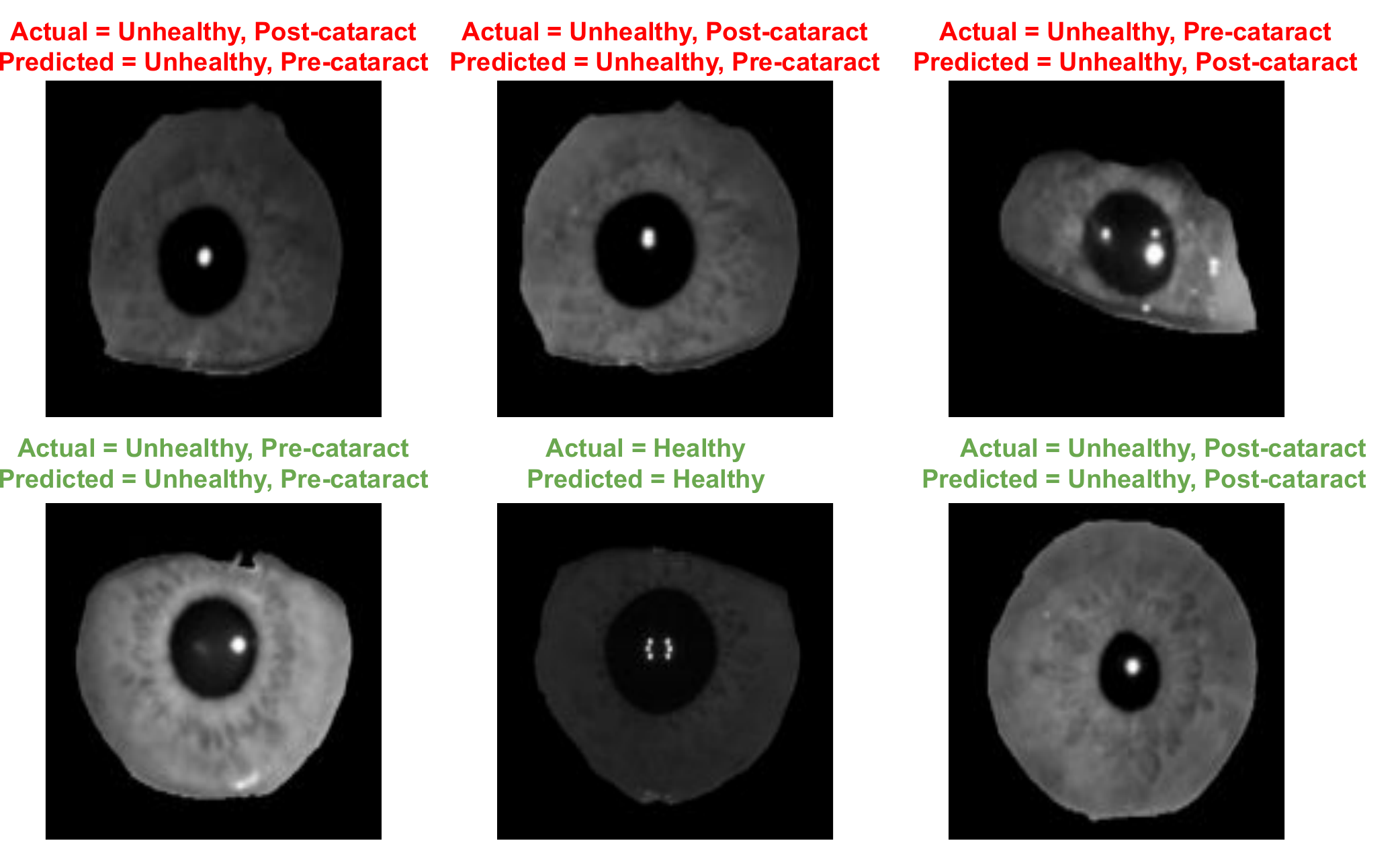}
\caption{Shows some correctly classified and misclassified samples from the dataset (best viewed in color).}
\label{fig:predictsamp}
\end{figure}

Fig.\ref{fig:tsne_plot} shows the tSNE plots of the healthy and unhealthy classes (Task 1), the first one is for the image space and second one is for the feature space. It is observed that the affected class (pre and post cataract) is well distinguishable from the healthy class. Fig. \ref{fig:predictsamp} shows the sample results of the proposed method. In the experiments, for Task 2, we have observed that some of the pre-cataract and post-cataract samples are misclassified with each other (as shown in Table \ref{tab:CM}).

\begin{table}[!h]
\centering
\caption{Illustrates the confusion matrix for two tasks}

\label{tab:CM}

\begin{tabular}{|c|l|c|c|c|}
\hline
\multicolumn{2}{|c|}{\multirow{3}{*}{Task 1}} & \begin{tabular}[c]{@{}c@{}}Predicted/\\ Actual\end{tabular} & Healthy & Unhealthy \\ \cline{3-5} 
\multicolumn{2}{|c|}{}                        & Healthy                                                     & 1.0     & 0.0       \\ \cline{3-5} 
\multicolumn{2}{|c|}{}                        & Unhealthy                                                   & 0.0     & 1.0       \\ \hline
\end{tabular}

\end{table}

\begin{table}[!h]
\centering
\begin{tabular}{|c|l|c|c|c|c|}
\hline
\multicolumn{2}{|c|}{\multirow{4}{*}{Task 2}} & \begin{tabular}[c]{@{}c@{}}Predicted/\\ Actual\end{tabular} & \begin{tabular}[c]{@{}c@{}}Pre-\\ Cataract\end{tabular} & \begin{tabular}[c]{@{}c@{}}Post-\\ Cataract\end{tabular} & Others \\ \cline{3-6} 
\multicolumn{2}{|c|}{}                        & \begin{tabular}[c]{@{}c@{}}Pre-\\ Cataract\end{tabular}     & 0.96                                                    & 0.4                                                      & 0.0    \\ \cline{3-6} 
\multicolumn{2}{|c|}{}                        & \begin{tabular}[c]{@{}c@{}}Post-\\ Cataract\end{tabular}    & 0.6                                                    & 0.94                                                     & 0.0    \\ \cline{3-6} 
\multicolumn{2}{|c|}{}                        & Others                                                      & 0.0                                                     & 0.0                                                      & 1.0    \\ \hline
\end{tabular}

\end{table}

\begin{comment}
\begin{table}[]
\caption{Illustrates the confusion matrix for two tasks}
\label{tab:CM}

\begin{tabular}{ccccc}
\centering
\hline
\textbf{Tasks} & \textbf{Classes}                                                              & \textbf{Actual}                                         & \textbf{Predicted}                                    & \textbf{\begin{tabular}[c]{@{}c@{}}Mis-\\ classified\end{tabular}} \\ \hline
Task 1         & \begin{tabular}[c]{@{}c@{}}Unhealthy\\ Healthy\end{tabular}                   & \begin{tabular}[c]{@{}c@{}}200\\ 100\end{tabular}       & \begin{tabular}[c]{@{}c@{}}200\\ 100\end{tabular}     & \begin{tabular}[c]{@{}c@{}}0\\ 0\end{tabular}                      \\ \hline
Task 2         & \begin{tabular}[c]{@{}c@{}}Pre-Cataract\\ Post-Cataract\\ Others\end{tabular} & \begin{tabular}[c]{@{}c@{}}100\\ 100\\ 100\end{tabular} & \begin{tabular}[c]{@{}c@{}}96\\ 94\\ 100\end{tabular} & \begin{tabular}[c]{@{}c@{}}4 as post\\ 6 as pre\\ 0\end{tabular}                  \\ \hline
\end{tabular}

\end{table}
\end{comment}
\begin{table}[!t]
\centering
%\vspace{0.5cm}
\caption{Shows the F1 scores obtained by varying the pre-trained models on the two tasks T1 and T2.}
\label{tab:basemodel_cmp}
\begin{tabular}{ccccc}
\hline
\multicolumn{1}{l}{\textbf{\begin{tabular}[c]{@{}l@{}}Model\end{tabular}}} & \multicolumn{1}{l}{\textbf{VGG16}}             & \multicolumn{1}{l}{\textbf{IV3}} & \multicolumn{1}{l}{\textbf{DN121}}         & \multicolumn{1}{l}{\textbf{RN50}}                     \\ \hline
\textbf{T1}                                                                                  & \begin{tabular}[c]{@{}c@{}}1.0 \end{tabular}         & \begin{tabular}[c]{@{}c@{}}0.96 \end{tabular}     & \begin{tabular}[c]{@{}c@{}}0.99 \end{tabular}         & \begin{tabular}[c]{@{}c@{}}1.0 \end{tabular}                  \\ \hline
\textbf{T2}                                                                                  & \begin{tabular}[c]{@{}c@{}}0.94 \end{tabular} & \begin{tabular}[c]{@{}c@{}}0.84 \end{tabular} & \begin{tabular}[c]{@{}c@{}}0.91 \end{tabular} & \textbf{\begin{tabular}[c]{@{}c@{}}0.97 \end{tabular}} \\ \hline

\end{tabular}

\end{table}

We next analyze the effect of base model, learning rate, and number of epochs:

\noindent \textbf{Effect of changing the base model:} For cataract classification, the performance of different deep learning models, viz. InceptionV3 (IV3) \citep{szegedy2016rethinking}, VGG16 \citep{simonyan2014very}, ResNet50 (RN50), and DenseNet121 (DN121)  are compared. As reported in Table \ref{tab:basemodel_cmp}, ResNet50 outperforms all other architectures for both the tasks and is an effective choice as a base model.

\begin{comment}
\begin{figure}[!t]
\centering
\includegraphics[width=0.4\textwidth]{fig/ROCeps100aug3class.png}
\caption{ROC curve for Task 2}
\label{fig:ROC3}
\end{figure}

\end{comment}

%\vspace{6pt}
\noindent \textbf{Changing the learning rate:} In this experiment, the learning rate is varied from 0.01 to 0.000001. We observe that the learning rate of 0.00001 outperforms the others yielding $100\% $ test accuracy for Task 1 and $96.67\% $ accuracy for Task 2. 

%\vspace{6pt}
\noindent \textbf{Changing the number of epochs:} We have also evaluated the performance by varying the number of epochs, and reported the results. It is shown that 100 epochs with learning rate = 0.00001 yields the best results for this classification problem. If we increase the number of epochs by 20, the results remain the same, beyond which the model starts overfitting. 
\section{Conclusion}
Cataract is a primary cause of visual impairment worldwide and cataract surgery is the most common elective surgical intervention. Typically, the prognosis, regular monitoring, and the decision of whether a patient should be taken up for surgery mostly depends on the discretion of the ophthalmologist. In resource constrained settings with limited experts, it is very important to have a clinical decision-support technique to improve sensitivity and specificity of cataract detection and monitoring. This paper presents a deep learning pipeline for cataract detection. To the best of our knowledge, this is the first work which proposes to use near infrared eye images, popularly used in iris biometrics, for cataract detection. A deep learning-based architecture, $PyramidNet$, is proposed for segmenting iris and pupil boundaries where the model fuses the coarse and fine information extracted from convolution blocks at different levels in a pyramid-like fashion. The segmented iris and pupil regions are then used for cataract classification via a multi-task network. Experiments performed on the cataract dataset shows that (i) effective cataract detection is possible in NIR domain, (ii) the proposed segmentation algorithm is effective in detecting iris and pupil boundaries even with challenging scenarios, and (iii) the overall cataract detection performance encourages such an approach to be used in automated decision support system. It is our assertion that the findings of this research and the availability of our datasets, will spur further research in this domain.

\bibliographystyle{model2-names}
\bibliography{refs}

\begin{comment}
\begin{subfigure}{.33\textwidth}
  \centering
  \includegraphics[width=.99\linewidth]{fig/Alcohol.pdf}
  \caption{ROC on IIITD Alcohol dataset}
  \label{fig:sfig2}
\end{subfigure}%
\begin{subfigure}{.33\textwidth}
  \centering
  \includegraphics[width=.99\linewidth]{fig/Dilation.pdf}
  \caption{ROC on XYZ Pupil Dilation dataset}
  \label{fig:sfig3}
\end{subfigure}

\begin{table}[]
\centering
\caption{Illustrating the sensitivity and specificity for Healthy vs. Unhealthy classes (Task 1) The proposed approach resulted in an improvement of 0.3 and 0.4 in sensitivity and specificity, respectively}
\label{tab:my-table}
\begin{tabular}{lcc}
\hline
Task 1            & \multicolumn{1}{l}{Sensitivity} & \multicolumn{1}{l}{Specificity} \\ \hline
Baseline          & 0.97                            & 0.96                            \\ \hline
\textbf{Proposed} & \textbf{1}                      & \textbf{1}                      \\ \hline
\end{tabular}
\label{fig:spec}
\end{table}

\end{comment}

\end{document}